\documentclass[12pt,a4paper]{article}
\usepackage{booktabs}
\usepackage{graphicx}
\usepackage{pict2e}
\usepackage{amsmath}
\usepackage{color}
\usepackage{epstopdf}
\usepackage{mdwlist}
\usepackage{amsfonts}
\usepackage{amssymb}
\usepackage{amsmath,rotating}
\usepackage{authblk}
\usepackage{geometry}
\usepackage{setspace}
\newtheorem{Theorem}{Theorem}
\newtheorem{Corollary}{Corollary}

\newcommand{\R}{\mathbb{R}}
\newcommand{\wv}{\mathbf{w}}
\newcommand{\ev}{\mathbf{e}}
\newcommand{\vv}{\mathbf{v}}
\newcommand{\xv}{\mathbf{x}}
\newcommand{\zv}{\mathbf{z}}

\newcommand{\qed}{\hfill \mbox{\raggedright \rule{.07in}{.1in}}}

\begin{document}

\title{Longitudinal Support Vector Machines for High Dimensional Time Series}

\author[1]{Kristiaan~Pelckmans}
\author[2,3,*]{Hong-Li Zeng}
\affil[1]{Department of Information Technology, Uppsala University, Sweden}
\affil[2]{Department of Mathematics, Uppsala University, Sweden}
\affil[3]{School of Science, and New Energy Technology Engineering Laboratory of Jiangsu Province , Nanjing University of Posts and Telecommunications, Nanjing 210023, China}
\date{}
\maketitle


\let\thefootnote\relax\footnote{*Corresponding Author: Hong-Li Zeng (Email: hlzeng@njupt.edu.cn).}


\begin{abstract}
	This paper considers the problem of learning a classifier from observed functional data.
	Here, each data-point takes the form of a single time-series and contains numerous features.
	Assuming that each such series comes with a binary label, the problem of learning to predict the label of a new coming time-series is considered.
	Hereto, the notion of {\em margin} underlying the classical support vector machine is extended to the continuous version for such data.
	The longitudinal support vector machine is also a convex optimization problem and its dual form is derived as well.
	Empirical results for specified cases with significance tests indicate the efficacy of this innovative algorithm for analyzing such long-term multivariate data.
\end{abstract}


\section{Introduction}

Longitudinal functional data consists of time-series rather than of single samples.
Typical examples are found in growth curves, in signals emanating from different celestial objects,
or in the analyses of medical trials where patients are followed for a period of time, see e.g. \cite{diggle1990time,diggle2002analysis}.
The traditional way to analyze this data is by inferring a stochastic model explaining the data,
making use of generalized linear models or random effect models.
However, such approaches are typically constrained by a curse of dimensionality \cite{hughes1968mean,oommen2008objective},
yielding difficulties when presented with high-dimensional data as for example resulting from image processing.

This paper explores a novel technique to deal with such time-evolved data,
by following the lines of thinking it as set out in the enormously successful literature on Support Vector Machines (SVMs)
\cite{cortes1995support,vapnik1998statistical,vapnik1999overview}.
The power of SVMs is generally attributed to three different foundations:
(i) the use of the well-understood device of convex optimization from both computational and theoretical point of view,
(ii) construction of nonlinear models by using represented kernels and high-dimensional linear approaches,
(iii) a solid learning theory.
We will focus in this paper on (i) and (iii), since nonlinear extensions - although straightforward - are not relevant for the application in mind.
The key point here is the notion of the {\em margin} for the separating hyperplane,
which is the SVMs try to find out and to separate the data into classes in an optimal way.

The goal of this analysis differs from that of statistical analysis
for longitudinal data,
which has been pointed out in the introduction of \cite{diggle2002analysis}.
Traditionally, one aims for the followings by the statistical analysis
\begin{itemize}
\item The impact of the deterministic covariates on the responses.
\item The presence and relevance of random effects underlying the observations.
\item The form of the autocovariance function underlying the responses.
\end{itemize}
Here however, the aim is slightly different:
{\em the objective is to find those features which evolve mostly in different ways in-between the two classes}.
This is known as {\em longitudinal data classification} \cite{marsallbaron00},
or {\em functional discriminant analysis} \cite{jameshastie2001,ramsaysilverman2005}.
Marshall et al. in \cite{marsallbaron00} describe a linear/quadratic discriminant analysis (LDA/QDA) for functional data.
The means of the classes, as well as they covariance matrices that determine the class labels are estimated by an iteration process.
James and Hastie proposed a functional linear discriminant analysis (FLDA) \cite{jameshastie2001} algorithm by extending the LDA method aiming at dealing with the irregular curves.
Here, there are no explicit covariates,
but the longitudinal data itself plays this role.
The response variable are in general multivariate, and one seeks to find the feature of those responses which differs at most in-between the two classes.
This aim goes much along the lines as taken by SVMs for classification, and it is exactly this correspondence that this paper explores in some detail.

A handful of authors have considered the analysis of longitudinal data with SVM-like methods before.
For example,  Suykens et al. presented a least square support vector machine \cite{suykens1999least},
which is a modified version of SVM that contains the least squares cost function and equality instead of inequality constraints.
Luts et al. extended the LS-SVM to a mixed effects LS-SVM model \cite{luts2012mixedLSSVM} aiming at classifying longitudinal data.
Chen and Brown \cite{chen2011lsvc} developed a Longitudinal Support Vector Classifier (LSVC) which extends the binary SVM
to deal with functional high dimensional dataset, basically by stacking the time-series in one large {\em input} vector.
Such approach is essentially bound to regularly sampled time-series.
Biau et al.  \cite{biau2005functional} give a theoretical account of classification with longitudinal data using a nearest neighbour approach.
Methods of machine learning have been widely applied to image processing, notably for medical imaging.
Wernick et al. surveys the field in \cite{Wernick2010MLimage}.

The approach presented here is tailored to the case of longitudinal analysis where
the different subjects originate from a common pool. That is, in the beginning of the experiment,
there is little hope that the different classes can be separated. Further up in the experiment  however,
it is reasonable to expect that the different classes present a growing difference.
Secondly, this approach is tailored towards the case where the different subjects are sampled at different time instances (`irregular sampling').
Note that in the case where all subjects are sampled at shared times (`regular sampling'),
one can reduce this application to a standard classification task, as done in \cite{chen2011lsvc}.
In order to cope with irregular sampling, the idea of a (parametric) margin function is introduced.
That is, samples and their labels are contrasted to the margin function evaluated at the relevant time instances.
This work focusses on the use of linear margin functions since they give a good trade-off between mathematical convenience and usefulness in the application mind.
Besides irregular sampling, margin function and tailoring to the evolutionary setup, this paper derives a convex program which is used to compute efficiently the solution.

The application motivating the development of this method goes as follows.
The aim is to classify breeding lines of an artificial selection experiment.
The brains of the individuals in those breeding lines are scanned, and the LSVM is used to detect differences in those for different selection regimes.
In a prototypical case, two classes of breeding lines are considered.
The first class evolves according to `{\em no selection}' and serves as a control.
The second class imposes a selection bias in the evolution process, which models `{\em artificial selection}'.
Now, we are interested how this difference in selection process manifests itself in the morphology of the brain.
This is, this is an instance of so called {\em artificial selection} or {\em selection-response} analysis \cite{falconer1996introduction}.
Hereto, we aim to find a classifier which separates both classes as well as possible under the evolutionary setup.
That is, in the beginning of the processes the two classes are overlapping.
They gradually diverge only when time (evolution) goes on.

This paper is organized as follows.
The next section details the proposed formulation.
Section III provides details on the margin function.
Section IV exemplifies the approach on artificially created case studies.
Section V concludes this paper.

\section{The Longitudinal Support Vector Machine (LSVM)}

This section gives the proposed learning scheme.
The following setup is adopted.
Let $X_i(t)$ denote a characterization of the $i$th subject at time $t$.
In the remainder of this paper, $X_i(t)$ is a random variable which takes values in $\R^p$ with $p>0$.
The $i=1, \dots, n$ subjects are divided into two classes, hereto we reindex the subjects such that
$X_i^+(t)$ belong to the first class for $i=1, \dots, n_+$, and $X_i^-(t)$ to the second one for $i=1, \dots, n_-$ so that $n=n_++ n_-$.
Equivalently, let $y_i=\{-1,1\}$ the label corresponding to $X_i$.
Then the goal is to find vectors $\wv_+,\wv_-\in\R^p$  such that
\begin{equation}
	\begin{cases}
		\wv^T_+ X_i^+(t)  + b \geq 0 & \forall i=1, \dots, n_+\\
		\wv^T_- X_i^-(t)  + b  < 0 & \forall i=1, \dots, n_-.
	\end{cases}
	\label{eq.separating}
\end{equation}
In case such a vectors exists, there are in general multiple solutions possible.
The key idea is to single one solution out.
Therefore, an additional principle is needed.

Let us start with the cross-sectional case, i.e. the case for fixed $t$.
Fix $t\geq 0$ and consider the two sets $\{X_i^+(t)\}_i$ and $\{X_i^-(t)\}_i$.
Then the {\em maximal margin} classifier or the hard SVM solves
\begin{multline}
	(\hat\wv_t, \hat{m}_t) = \arg\max_{\wv_t,m_t} m_t
	\\
	 \mbox{ \ s.t. \ }
	\begin{cases}
		\wv_t^TX_i^+(t) \geq  m_t & \forall i=1,..,n_+ \\
		\wv_t^TX_i^-(t) \leq - m_t & \forall i=1,...,n_-	 \\
		\|\wv_t\|_2=1.
	\end{cases}
	\label{eq.mm0}
\end{multline}
Here, $m_t$ denotes the size of the margin at time $t$.
This problem can be converted into a convex Quadratic Program (QP) using a change of variables.

Now, this can be done for any $t$, while it is natural to impose to have the same vector $\wv_t=\wv$.
The question is of what to do with the margins $\{m_t\}_t$.
Rather than having independent constants $m_t$ for every $t$, a parametric model of the
{\em margin function} is imposed
\begin{equation}
	m_t = a t + d,
	\label{eq.marginfunction}
\end{equation}
with parameters $(a,d)$.
Note that this function can go negative, resulting in a negative margin:
this means that samples from both classes can take the same values.
This is a natural feat when the two classes have no clear distinction:
this typically happens in the begin phase of the longitudinal study when
the differentiating process has only started.

One very attractive feature of this formulation is that this approach can cope naturally
with irregularly sampled data. That is, every time-instant that a sample is available, it is
checked for with the margin function. Formally, let
each subject $X_i$ for $i=1, \dots, n$ be sampled $n_i$ times $t_{i1}\leq \dots \leq t_{in_i}$.
The resulting classifier becomes
\begin{multline}
	(\hat\wv, \hat{a}, \hat{d}) = \arg\max_{\wv, a, d}  \sum_{i=1}^n \sum_{j=1}^{n_i} (at_{ij}+d)
	\\
	\mbox{ \ s.t. \ }
	\begin{cases}
		y_i(\wv^TX_i(t_{ij}) +b) \geq a t_{ij} + d & \\
		\ \ \ \ \ \  \forall i=1, \dots n, \ \forall j=1, \dots, n_i \\
		\|\wv\|_2=1,
	\end{cases}
	\label{eq.mm}
\end{multline}
where the label of a new observation $X_\ast$ which is measured
at time points $t_1 \leq \dots t_{n_\ast}$, is predicted as
\begin{equation}
	\begin{cases}
		y_\ast = +1 & \mbox{ \ if \ } \sum_{j=1}^{n_\ast}(\wv^T X_\ast(t_j)+b) \geq 0 \\
		y_\ast = -1 & \mbox{ \ if \ } \sum_{j=1}^{n_\ast}(\wv^T X_\ast(t_j) +b)  \leq 0\\
	\end{cases}
	\label{eq.rule}
\end{equation}
This rule is not always conclusive:  there might exist points that satisfy both
margins simultaneously (when the margin function is negative),
and there are points satisfying neither (when the margin is strictly positive).
Note that this formulation aggregates the margins by using $\sum_{j=1}^{n_t}$.
While this choice is ad-hoc at this point, this {\em average margins} idea comes with
desirable properties as shown later.

Observe that $\wv$ will always be taken as large as possible, such that one can safely
replace the constrain $\|\wv\|_2= 1$ by $\|\wv\|_2\leq 1$.
However, there is a slight problem when the overall average margins is at best negative, then
a $\|\wv\|_2=0$ would be favoured. This is to be checked case-by-case.
Secondly, we introduce slack variable
$\{\epsilon_{ij}\}_{ij}$, making it possible to violate the margin in a small number of cases,
increasing robustness of the algorithm.
This results in the following convex QP:
\begin{multline}
	\max_{\wv, a, b,d,\{\epsilon_{ij}\}_{ij}}  \sum_{i=1}^n \sum_{j=1}^{n_i} (at_{ij}+d)
	- C \sum_{i=1}^n \sum_{j=1}^{n_i} \epsilon_{ij}
	\\
	\mbox{ \ s.t. \ }
	\begin{cases}
		y_i(\wv^TX_i(t_{ij})+b) \geq a t_{ij} +d -\epsilon_{ij}& \\
		\ \ \ \ \ \  \forall i=1, \dots n, \ \forall j=1, \dots, n_i \\
		\epsilon_{ij} \geq 0 & \\
		\ \ \ \ \ \  \forall i=1, \dots n, \ \forall j=1, \dots, n_i \\
		\|\wv\|_2\leq 1.
	\end{cases}
	\label{eq.mm2}
\end{multline}
where $C\geq 0$ is a user-defined (tuning) parameter.
The formulation eq. (\ref{eq.mm2}) is {\em convex}, and can be solved efficiently using polynomial time solvers for
convex Quadratically Constraint Linear Programs (QCLPs). Appendix A derives the dual of this problem,
which is computationally much more attractive to solve in case of large $p$.

It is useful to note how this formulation copes with the themes traditionally addressed
in a statistical analysis of such data.
First of all, the LSVM only aims to identify the difference in drift amongst the two classes.
That is, we do not aim to {\em model} the dynamics underlying the individual observation,
and there is no need to recover the {\em autocovariance functions}.
This would be too difficult a task anyways for applications where $p\gg n$,
since the covariance function would be of order $p^2$.

Secondly, this formulation deals with {\em random effects} of the $n$ different subjects as follows.
The inequality of the margin merely takes into account the {\em worst case} at any given time-point.
(In case of $C<\infty$, one relaxes effectively to {\em approximatively} worst case).
That is, random effects are taken into account  by using `$\leq$' rather than `$=$' in the constraints of (\ref{eq.mm2}).
Again, there is no intension to recover the random effects, only to infer a global (i.e. population) property differentiating the two classes.

\section{The Margin Function}

This section addresses the question why the `average margins' is a good idea.
At first, it translates the intuition as to how the two classes drift apart in the envisioned application.
But there is also a more formal reason:
there exist in general only a small number of classifiers having a fixed `average margin'.
This is made formal using the device of shattering numbers and VC dimension, as discussed in \cite{vapnik1999overview, sabato2013distribution}.

Recall a main result for (traditional) classifiers.
Let $\{\zv_1, \dots, \zv_m\}\subset\R^d$ be a set of vectors
and let $\{f(\cdot; \wv)=I(\zv_i^T\wv \geq 0): \wv\in\R^d\}$ be a class of functions.
Consider the set
\begin{equation}
	\mathcal{F}_m =  \left\{\left(f(\zv_1;\wv), \dots,  f(\zv_m;\wv)\right)\in\{0,1\}^m, \wv: \  \|\wv\|_2\leq 1\right\}.
	\label{eq.F}
\end{equation}
The size of this set is bounded by its {\em shattering} number
\begin{equation}
	\Pi\mathcal{F}_m = \sup_{\zv_1, \dots, \zv_m} |\mathcal{F}_m|.
	\label{eq.F}
\end{equation}
This is in turn bounded by Sauer's Lemma, as follows
\begin{equation}
	\Pi\mathcal{F}_m \leq \sum_{k=0}^{h} {m \choose k}.
	\label{eq.F}
\end{equation}
Here $h$ is the so-called VC-dimension of the set $\mathcal{F}_m$.
This one is to be finite for growing $m$ (necessary and sufficient condition)
in order for an algorithm to {\em learn} from data.

For margin-based classifiers, the following result is paramount.
\begin{Theorem}
	Given a set of samples $\{\zv_1, \dots, \zv_m\}\subset\R^p$ of size $m$,
	such that  $\|\zv_i\|_2\leq r<\infty$.
	Let $\mathcal{F}$ be a set of functions with a minimal margin $d$,
	and consider their projections on the sample as
	\begin{equation}
		\mathcal{F}_m =  \left\{\left(f(\zv_1), \dots,  f(\zv_m)\right): \forall f\in\mathcal{F}\right\}.
		\label{eq.F}
	\end{equation}
	Assuming that every $f$ separates the samples by a margin of size at least $d>0$,
	then this set has VC dimension bounded by
	\begin{equation}
		h \leq \min\left(m+1, \left\lceil\frac{r^2}{d^2}\right\rceil\right),
		\label{radiusmargin}
	\end{equation}	
	If $d\leq 0$, the VC-dimension is bounded by $m+1$.
\end{Theorem}
This is the learning-theoretical foundation of the use and design of SVMs for classification,
fundamental in the work of Vapnik, see e.g. his IEEE overview paper \cite{vapnik1999overview}
and the references therein.
Recent advances in VC theory for large-margin classifiers are summarized and extended in
\cite{sabato2013distribution}.

The longitudinal problem is reduced to this case as follows.
For ease of notation, let $n_1= \dots n_m =n$
and fix constants $t_1\leq \dots \leq t_n$ at which times {\em all} $m$ subjects are measured,
that is $t_1=t_{11}=\dots = t_{m1}$, $t_i=\dots$ and $t_n = t_{1n}=\dots = t_{mn}$.
Hence the deterministic average-margin is given as
\begin{equation}
	m(a,d) = \frac{2a}{n}\sum_{j=1}^n t_j + 2d.
	\label{eq.sumofmargins}
\end{equation}
The decision functions are given as
\begin{equation}
	f(X; \wv) =
	I\left( \sum_{j=1}^n\wv^T X(t_j) \geq 0\right).
	\label{eq.sumofmargins2}
\end{equation}
So, one has that
\begin{equation}
	\mathcal{F}_m =  \left\{\left(f(X_1;\wv), \dots,  f(X_m;\wv)\right), \wv: \ \|\wv\|_2\leq 1\right\}.
	\label{eq.F}
\end{equation}
The {\em size} of this set is bounded when only considering functions realizing a certain average-margin:

\begin{Corollary}
	Let $t_j\leq \dots \leq t_n$ be fixed constants as defined before.
	Consider the functions $\mathcal{F}(\mu)$ realizing a minimal average-margin $\mu>0$ on the sample.
	The corresponding set $\mathcal{F}_m(\mu)$ has a VC dimension bounded by
	\begin{equation}
		h \leq  \min\left(m+1, \left\lceil\frac{n^2r^2}{\mu^2}\right\rceil\right),
		\label{eq.F}
	\end{equation}
	where $\mu\geq m(a,d)$ is defined as in (\ref{eq.sumofmargins})
	and $r\geq \max_{i,j} \|X_i(t_j)\|_2$  a.s..
\end{Corollary}
This follows immediately by the following reduction. Let
\begin{equation}
	\xv_i = \left( X_i(t_1)^T, \dots, X_i(t_n)^T \right)^T\in\R^{pn},
	\label{eq.xv}
\end{equation}
and
\begin{equation}
	\tilde{\wv} = \left( \wv^T, \dots, \wv^T \right)^T\in\R^{pn}.
	\label{eq.wwv}
\end{equation}
Then $I(\xv_i^T\tilde{\wv}\geq 0)$ is equivalent to (\ref{eq.sumofmargins2}),
and it is a classical, linear learning rule.
Moreover, it has margin $\frac{2}{n}\sum_j d_{t_j} \geq \frac{2}{n}\sum_j at_j + 2d$,
and radius bounded by $nr$.
Observe that the $\frac{1}{n}$ factor comes in as $\|\tilde\wv\|_2\leq n$ as $\|\wv\|_2\leq 1$.
hence the radius-margin bound of Theorem 1 applies directly.
Note also that if for each $t_j$, $h_j$ cases can be shattered with margin $b_j$ by $\{\wv^T(\cdot): \|\wv\|_2\leq 1 \}$,
one can also shatter with $\mathcal{F}_m$ at least $m\geq \min_j h_j$ subjects in the longitudinal case, with margin $\sum_j b_j$.

Hence, maximizing $m(a,d)$  is also motivated by a learning-theoretic perspective.
Furthermore, this bound gives useful insight in the algorithm.
Firstly, if the average margin cannot be made strictly positive, the algorithm doesn't {\em learn} from the $m$ (independently sampled) subjects.
Secondly, the average margin has to grow faster than $n$ in order for new time-samples $t_j$ to be useful.
Thirdly, the dimensionality $p$ of the problem does not influence its learning behaviour directly.
This is an important insight as it says that this approach is not hampered directly by the curse of dimensionality.
Note however that one has to be careful not to have $r$ grow unboundedly when having the dimension increase.

\section{Case Studies}

\subsection{{\em A Simple Case}}

The above method is applied to simple simulations where $p=36$.
This experiment abstracts the task of extracting relevant features of brain images (scans) of evolving species,
and how to separate between two different evolution strategies.
Think of {\em one subject} $X_i$ in the longitudinal setting as a single {\em (evolutionary) breeding line}.
That is, initially all individuals are sampled from a global pool.
We distinct between two classes: the breeding lines subject to {\em artificially selection}, and the {\em control} lines.
Each new individual is a slight mutation of its direct ancestors.
But lines of the first class have different selection strategies than the lines of the control class.
An example of each is given in Fig. \ref{fig:case}.

We applied the LSVM method to this simple case.
Here, $C$ is fixed to $0.001$ according to the training validation performance.
By solving the dual form of eq. (\ref{eq.mm2}), one obtains the matrix $\hat{\wv}$ as shown in Fig. \ref{fig:heatmap}(b),
which is comparable to the `true' preference $\wv_0$ in Fig. \ref{fig:heatmap}(a).

At $t=0$, $X_i(0)\sim \mathcal{N}(0_p,I)$ with $I$ the identity matrix of appropriate dimension. The mechanisms of {\em control} and {\em artificial selection} are implemented by the following transition rules
\begin{multline}
	\begin{cases}
		X_{i,t+1} = X_{i,t}  + \sigma U_{i,t}, \ \ ~~~~~~~~\text{control} & \\
		X_{i,t+1} = X_{i,t} + \sigma U_{i,t} + \ev , \ \ ~~\text{artificial selection} & 		
	\end{cases}
	\label{eq.rules}
\end{multline}
with parameter $\sigma$ indicates the level of the noise
$U_{i,t} \sim \mathcal{N}(0,1)$,
 $\ev=(2, 2, 0_{p-2})^T\in\R^{p}$ denotes the drift.
The vector $\wv_0\propto (1, 1, 0_{p-2})^T\in\R^{p}$ as shown in Fig. \ref{fig:heatmap}(a)
represents the {\em true preference} in the artificial selection protocol.

\setlength{\unitlength}{1.mm}
\begin{figure}[htbp] 
	\begin{center}
        \includegraphics[width=0.5\textwidth]{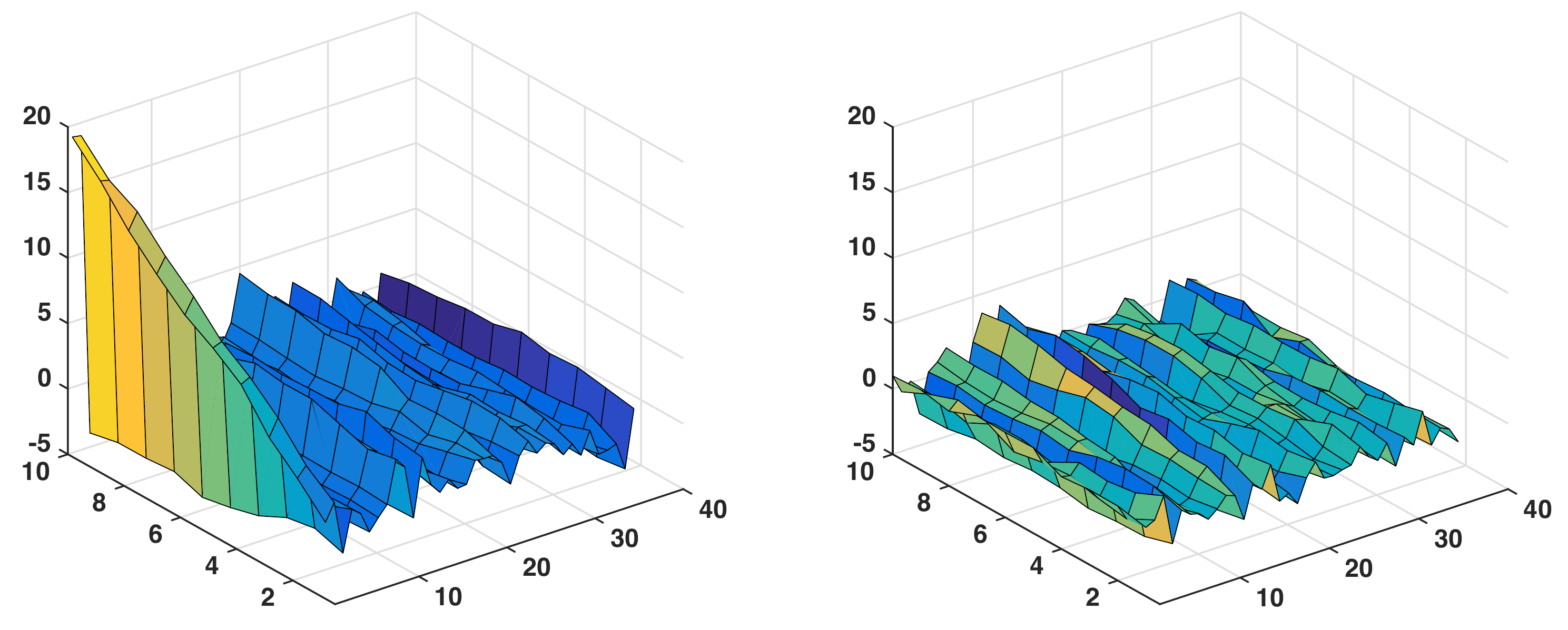}
	\end{center}
  \begin{picture}(0,0)
		\put(58,43){(a)}
		\put(100, 43){(b)}
		\put(36,23){\rotatebox{90}{\textbf{$X_i$}}}
		\put(78,23){\rotatebox{90}{\textbf{$X_j$}}}
		\put(110,9){\rotatebox{0}{\textbf{$p$}}}
		\put(88,10){\rotatebox{0}{\textbf{$t$}}}
		\put(70,9){\rotatebox{0}{\textbf{$p$}}}
		\put(45,11){\rotatebox{0}{\textbf{$t$}}}
   \end{picture}
   \caption{Two example lines of the simple example with $X_i(t)\in\R^{36}$ for $t=1,\dots, 10$.
    Left: an {\em artificial selection} breeding line $X_i$;
    Right: a {\em control} breeding line $X_j$.
    Here, $\sigma=0.5$.}
   \label{fig:case}
\end{figure}
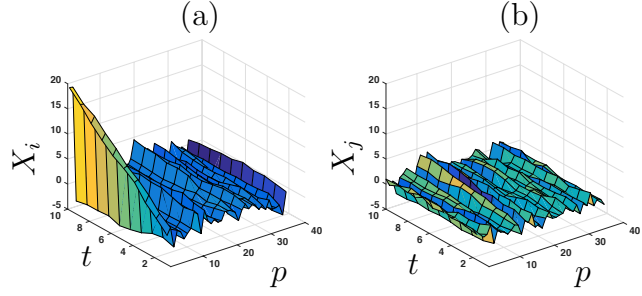

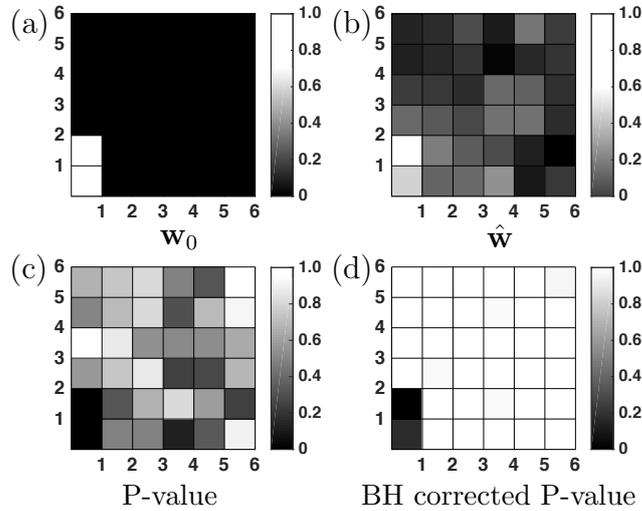
\begin{figure}[htbp] 
   	\begin{center}
		\includegraphics[width=0.5\textwidth]{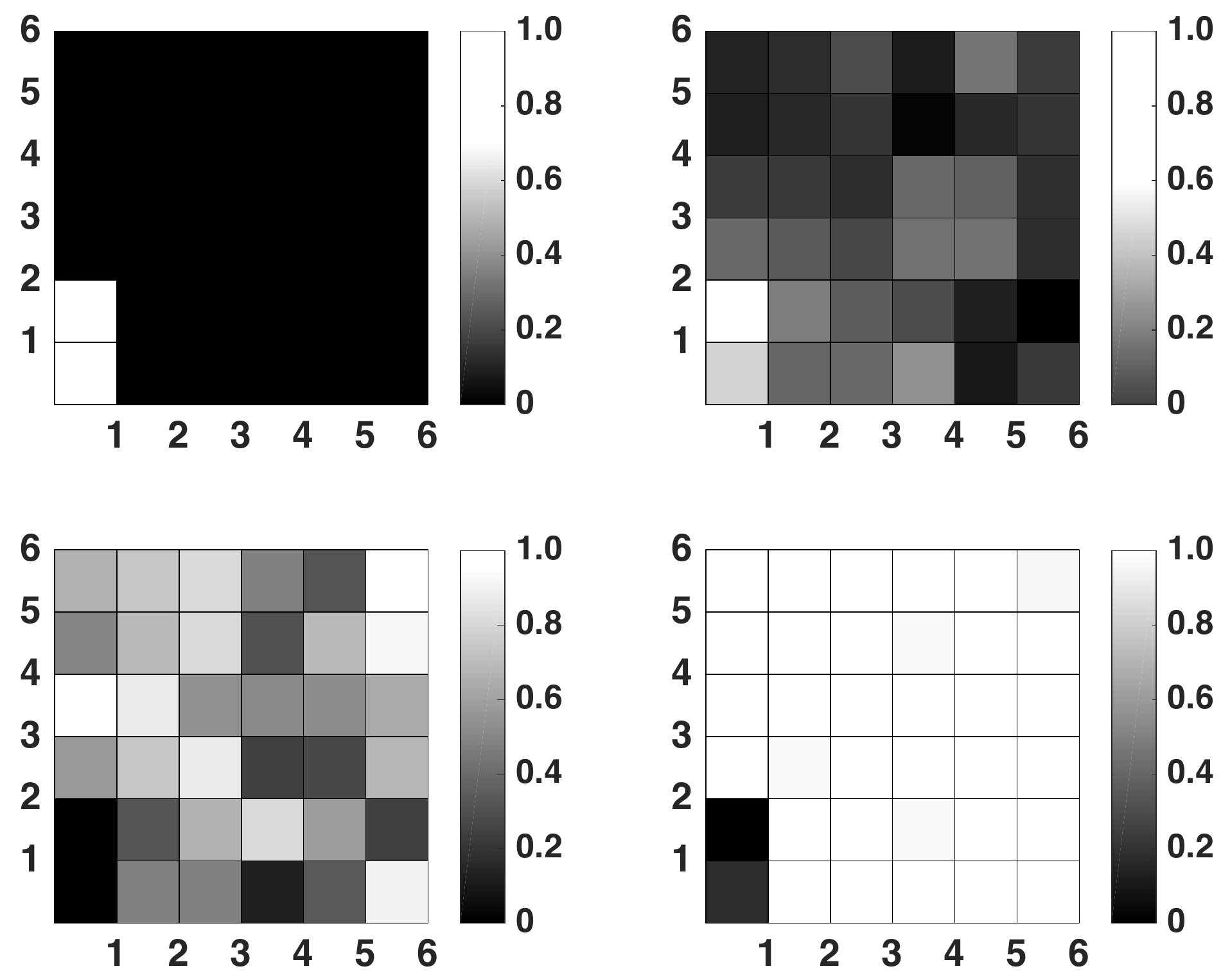}
	\end{center}
	\begin{picture}(0,0)
		\put(35.2,68.5){{(a)}}
		\put(77.5,68.5){{(b)}}
		\put(35.2, 36){{(c)}}
		\put(77.5,36){{(d)}}
		\put(55,40.5){\textbf{$\wv_0$}}
		\put(98,40.2){\textbf{$\hat{\wv}$}}
		\put(50,6){{\small{P-value}}}
		\put(81.5,6){{\small{BH corrected P-value}}}
  \end{picture}
    \caption{Representation of the recovered result.
   (a) `true' preference $\wv_0$;
   (b) values of $\wv$ as recovered by  by solving the dual form of eq. (\ref{eq.mm2});
   (c) P-value for each pixel by permutation test;
   (d) values for thresholding the adjusted P-value by the multiple testing method of BH.
   Here, $C=0.001$, $\sigma=0.5$, 10 replicates for each treatment, $t=10$ for each replicate, and using 10000 permutations for the test.}
   \label{fig:heatmap}
\end{figure}

\subsection{{\em Significance Test}}

Generally, a significant results is accepted when a P-value is less than a pre-fixed significance level $\alpha$ (say 0.05) \cite{fisher1925statistical, cumming2013understanding}.
The P-value is the probability of attaining results that are described by the {\em null hypothesis} $H_0$,
while $\alpha$ is the probability of rejecting $H_0$ given that it holds true.
Here, $H_0$ denotes for each pixel it is irrelevant (or that $\wv_i=0$).
Practically, the permutation test \cite{fisher1960design} is used to check if the inferred $\hat{\wv}$ by LSVM is significant.
This technique is widely used in medical image processing \cite{golland2003permutation, statistical_significancemapSVM}.
For images with high dimension, the inference of $\hat{\wv}$ is an instance of a so-called multiple testing problem  \cite{rupert2012simultaneous}.
One infers all elements of $\hat{\wv}$ simultaneously for all $p$ pixels.
Several techniques have been developed to correct the testing \cite{bonferroni1936teoria, benjamini2001control}.
Here, corrected P-values for each pixel $i$ are computed as in \cite{benjamini1995controlling}, referred to as BH.

For the simple case, the heat-map of P-values for each individual is presented in Fig. \ref{fig:heatmap}(c).
The P-values of the first two pixels is 0.0007 and 0.0001 respectively,
which are much smaller than that of the others.
To avoid rejecting the $H_0$ for some pixels wrongly, the adjusted P-values are calculated according to the BH method.
For the first two relevant pixels, they are 0.013 and 0.0036 respectively, as shown in black in Fig. \ref{fig:heatmap}(d).
That for other irrelevant ones are around 1, as shown in white in Fig. \ref{fig:heatmap}(d).

\subsection{{\em An Artificial Fish Brain Experiment}}
A more complicated case with two groups of augmented fish brains is studied.
One group is based on {\em artificial selection} while the other one for {\em control}.
The experiments found that the group difference increases over generation in relative brain size \cite{kotrschal2013artificial, kotrschal2015larger, kotrschal2015positive}.

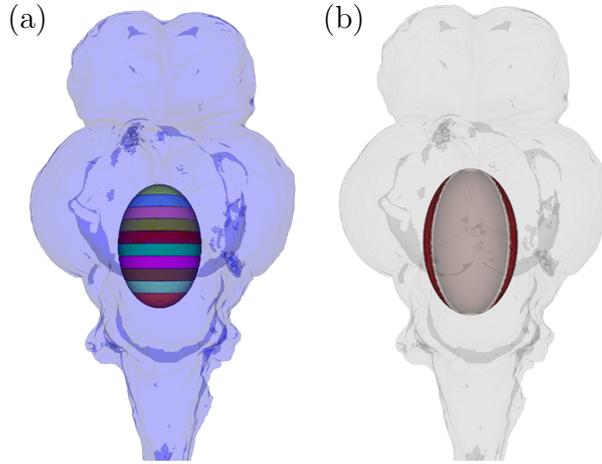
\begin{figure}[htbp] 
	\begin{center}  
		\includegraphics[width=0.245\textwidth]{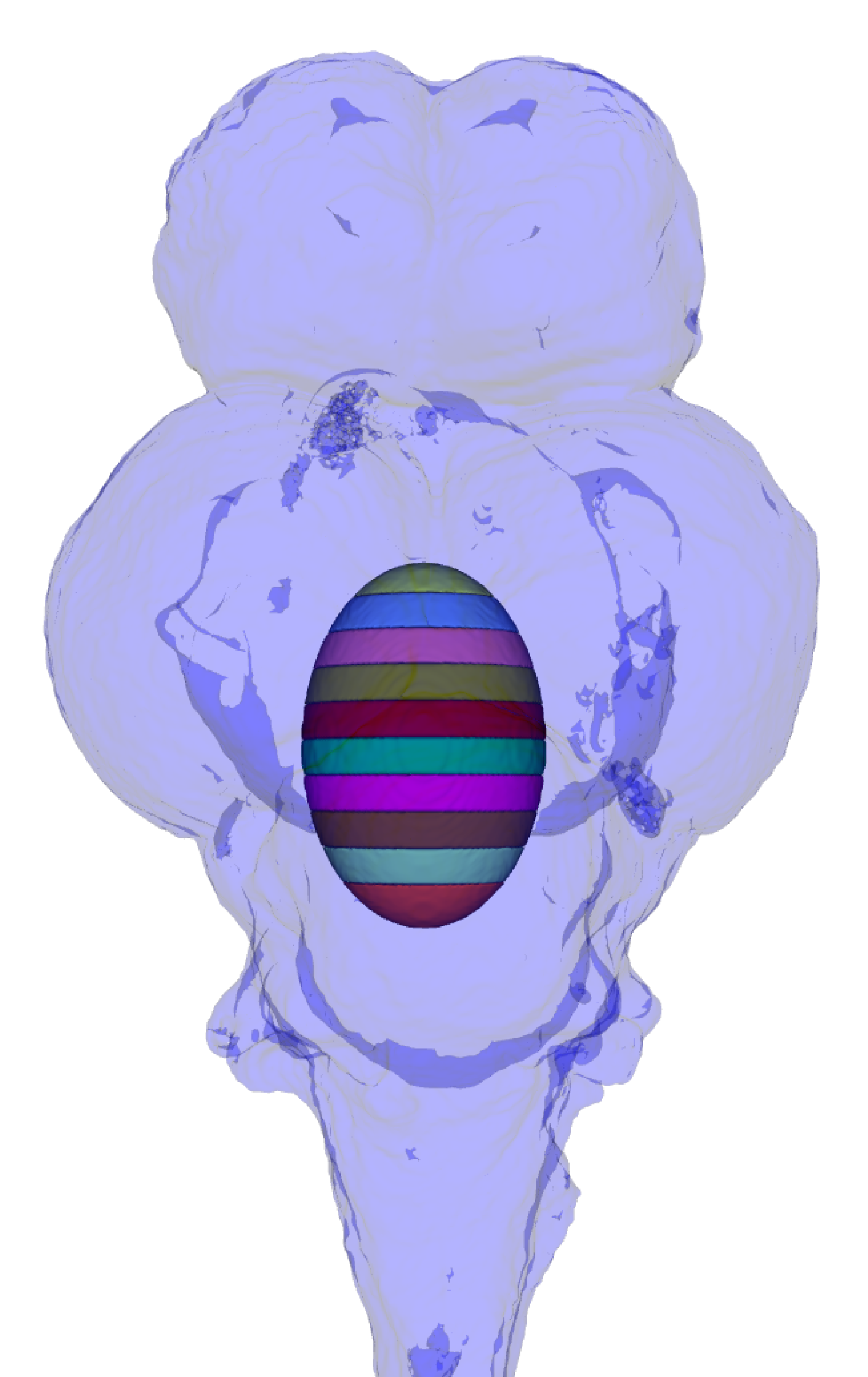}
		\includegraphics[width=0.25\textwidth]{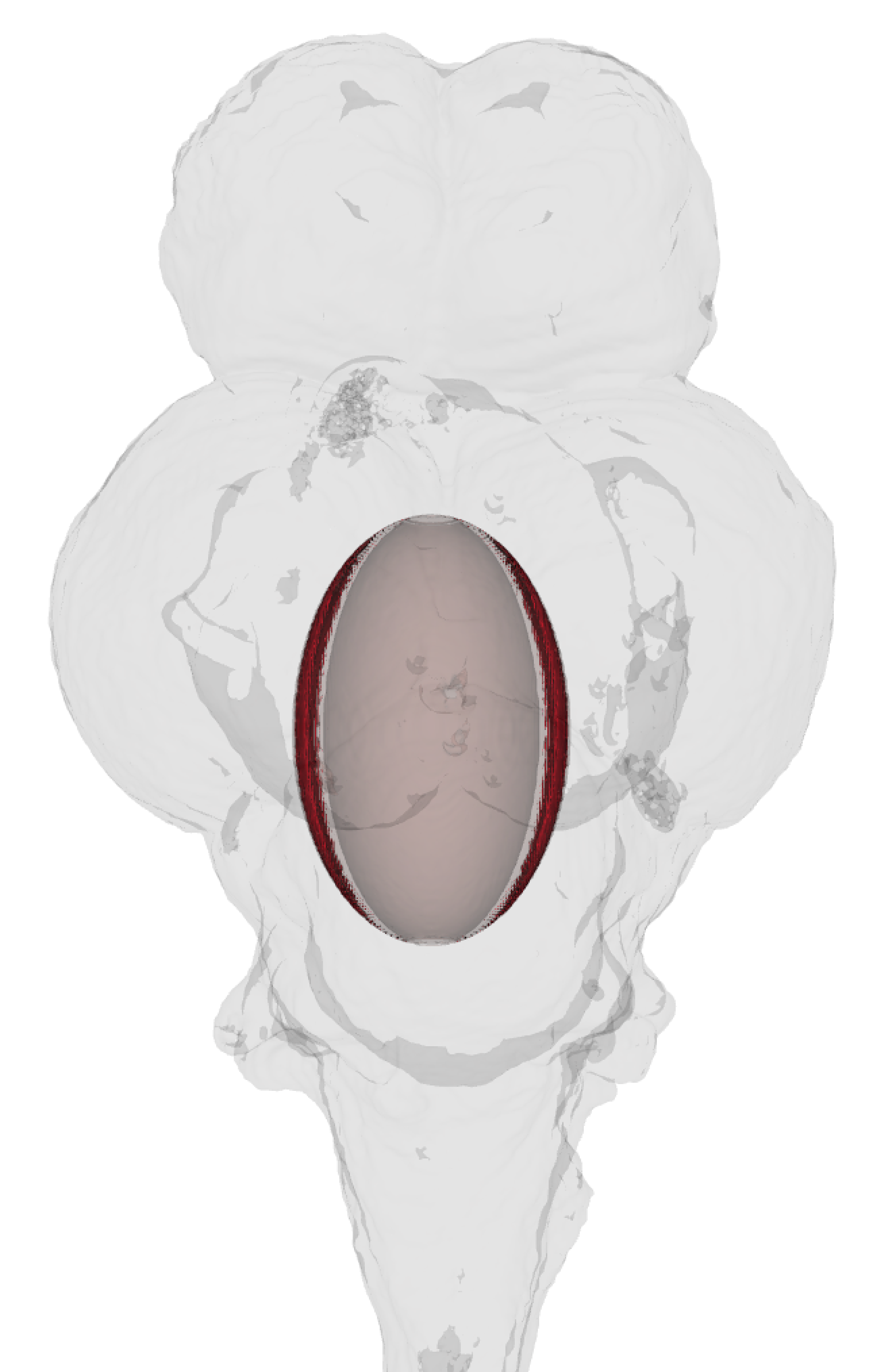}
	\end{center}
	\begin{picture}(0,0)
   \put(38.5,67){(a)}
   \put(80,67){(b)}
\end{picture}
   \caption{(a) An example of real fish brain scans.
    For the sake of visualization, the whole brain body (blue part) are shown with an opacity of 0.3 while 1 for the inside artefact;
   (b) the map of P-value for each voxel by permutation test,
   arranged in a 3D fish brain structure for visualization.
   The dark red voxels means their P-values $\leq 0.05$.
   This indicates the relevant parts from two groups of augmented brains.
   The light red voxels inside the dark red region have P-values (0.05, 1) while the grays for P-values=1 (not relevant).
  Parameters: $t=5$, $a_0$=0.04, $s_0^x=s_0^z=0.15$, $\lambda=0.01$.}
   \label{fig:fishbrain}
\end{figure}

The brains of dead fish were removed and stained for 2 days with Osmiumtetraoxide (1\% in PBS),
then washed and embedded in 3\% agar for the subsequent scanning.
The collected brains were scaned by microcomputed
tomography (SkyScan 1172, Bruker microCT, Kontich,
Belgium).
The CT scans are rearranged in 3D volumes.
An example of the scans is shown in Fig. \ref{fig:fishbrain}(a),
which contains around $10^7$ voxels.

To expore the efficiency of LSVM in detecting group differences,
 ellipsoid artefacts are introduced into a 3D volume of the brain scan.
The diameters of the augmented artefact varies over generations $t$ along X and Z directions in the simuation.
While the intensities of the artefacts are 2 to 20 higher than those at the same voxel locations.
They are calibrated along the Y-axis, as shown in the calibrated ellipose of Fig. \ref{fig:fishbrain}(a).

The $i$th line has an artefact of size $\max(0,s_i^x)$ along X-direction
and $\max(0,s_i^z)$ along Z-direction.
The transition rules for {\em control} and {\em artificial selection} along X and Z direction are as follows:
\begin{multline}
	\begin{cases}
		s_{i,t+1} = s_{i,t}  + \lambda U_{i,t}, \ \ ~~~~~~~~\text{control} & \\
		s_{i,t+1} = s_{i,t} + a_0 + \lambda U_{i,t}, \ \ ~~\text{artificial selection}
	\end{cases}
	\label{eq.3dfishbrain}
\end{multline}
where $U_{i,t}\sim\mathcal{N}(0,1)$,  $a_0>0$ the drift term in the artificial selection and
$\lambda$ the fluctuation level of $s_i$ in both cases over generations.

There are 4 breeding lines,
half with artificial selection and half for control.
For each breeding line, 5 generations are included.
From Fig. \ref{fig:fishbrain}, one finds that LSVM detected the consistently changing part (the `artefact').
The map of P-values for each voxel of this case is presented in Fig. \ref{fig:fishbrain}(b).
The P-values for the relevant part are close to 0 (the dark red voxels) while that for the other unchanged parts are almost 1 (light red and gray voxels).

\section{Discussion: The Performance of The Dual LSVM Algorithm}
\begin{figure}[htbp] 
	\begin{center}
		\includegraphics[width=0.5\textwidth]{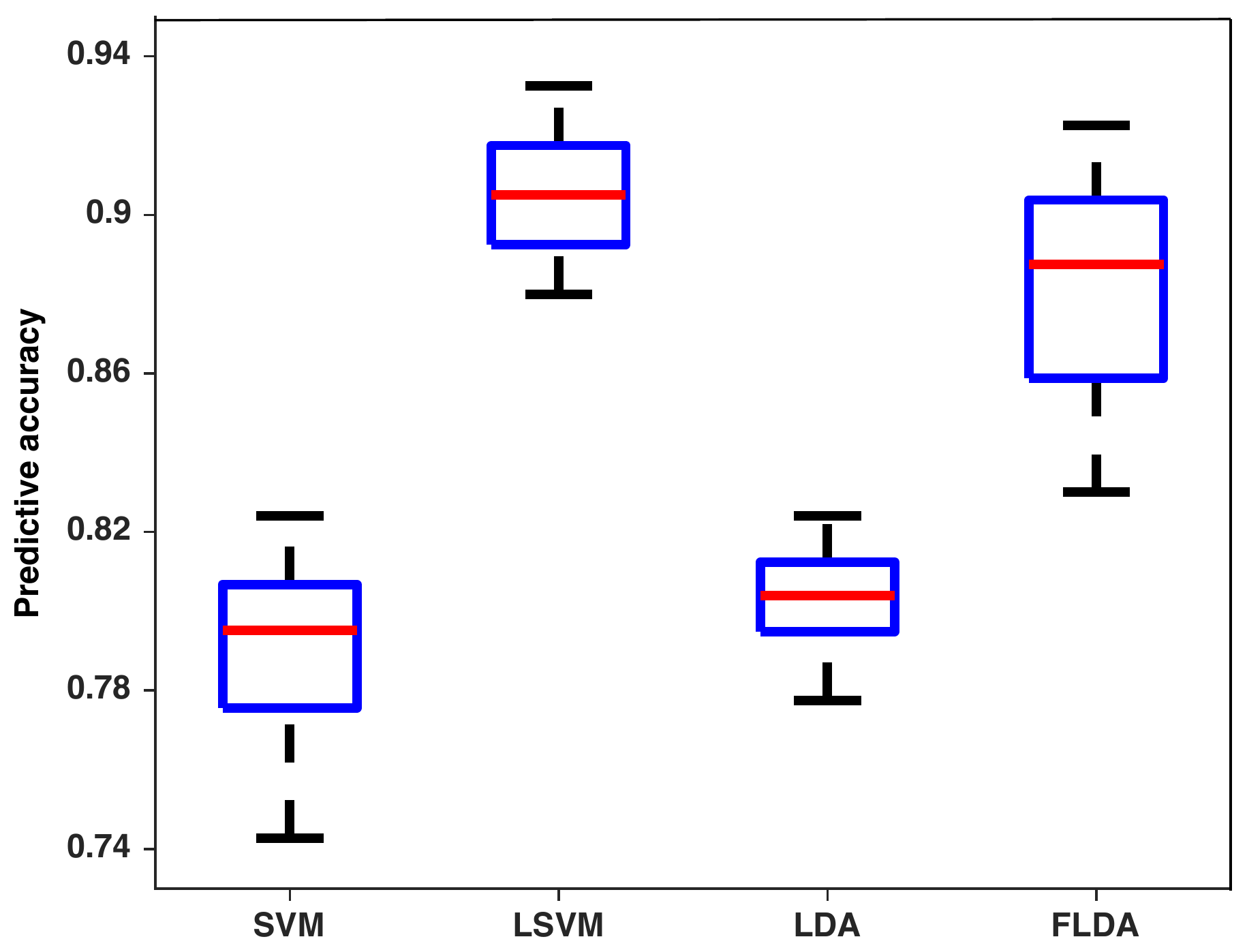}
	\end{center}
	\begin{picture}(0,0)
		 \put(51,50.5){\includegraphics[width=0.7in]{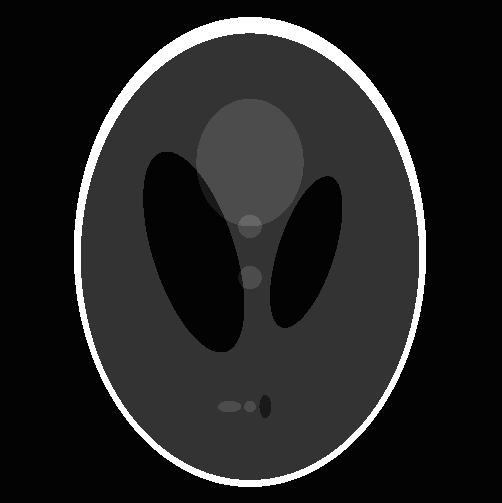}}
  	\end{picture}
   \caption{
   Main panel: Predictive accuracy rate for SVM, LSVM, LDA and FLDA algorithm on 2D Shepp-Logan (SL) phantom.
   Inset: an example of a 2D SL phantom.
   Parameters: $p =64\times64$,  10 replicates for artificial selection and control respectively composed the training dataset.
   2000 replicates for each treatment in the testing dataset,
   $t=10$. With 21 independent trials, P-value = 0.003 for LSVM versus FLDA.}
   \label{fig:boxplot}
\end{figure}

The efficacy of the LSVM is compared with the other three ones: Binary Support Vector Machine (SVM),
Linear Discriminant Analysis (LDA) and functional linear discriminant analysis (FLDA) as described in \cite{jameshastie2001}.
Performance of those 4 are compared using the classical Shepp-Logan (SL) phantom dataset, $p=64\times64$, as displayed in the inset of Fig. \ref{fig:boxplot}).
As before in eq. (\ref{eq.3dfishbrain}), the size of the ellipsoid (short axis, left black ellipsoid) is varied according to a simple rule.
As presented in the main panel of Fig. \ref{fig:boxplot}, the LSVM and FLDA obviously work better than their original versions.
Application of the sign-rank test for testing the difference between LSVM and FLDA gives a P-value of 0.3\%, indicating that LSVM performs significantly better than FLDA.
Since (F)LDA requires computation of the inverse of a $p\times p$-matrix, it is not suited well to handling large-dimensional tasks.

\section{Conclusion}
In this paper, we extend the classical SVM algorithm to classify the longitudinal data.
Instead of concatenating all images of a breeding line together, we introduced the parameterized margin.
As a direct consequence, irregular sampling can be handled straightforwardly.
The approach is tailored to the evolutionary setting where the breeding lines originate from a common pool.
The derivation results in a convex program which can be solved efficiently, while the dual derivation is used to handle high-dimensional problems proficiently.
Finally, we indicate a technique (permutation test, and corrections using BH) to convert the inference into P-values which are useful for statistical analysis.
Numerical studies indicate the efficiency and usefulness of this approach in an artificial evolutionary experiment.

Two important issues are pending:
(i) while we can handle large dimensions of $p=O(10^7)$, even larger dimensionalities are encountered in brain imaging.
This points to the use of methods of wavelets or related.
(ii) correct tuning of the value of $C$ is crucial for obtaining the desired performance.
It is a challenging question how to do this effectively in case no (artificial) validation sets are available.

\section*{Acknowledgment}
The authors would like to thank grant 2013.0072 founded by ``Knut and Alice Wallenberg Foundation" for the research on ``Social behaviour and brain".


\section*{Appendix A: Dual form of eq. \ref{eq.mm}}
The dual form of eq. \ref{eq.mm} can be derived by a variable transform as is done for the derivation of the standard SVM \cite{cortes1995support}.
We want to choose the new variables $\vv, a', b', d' $ such that
\begin{equation}
\frac{ \sum_{i,j}(at_{ij}+d)}{\sum_i n_i} \|\vv\|_2 = 1,
\end{equation}
Hence, maximizing the total margin $\left(\sum_{ij}(at_{ij} + d) \right)$ becomes equivalent to minimizing the norm $\|\vv\|_2$.
Then, the classifier becomes
\begin{multline}
	 \arg\min_{\vv, a'}  \frac{1}{2}\|\vv\|_2^2
	\\
	\mbox{ \ s.t. \ }
	\begin{cases}
		y_i(\vv^TX_i(t_{ij}) + b') \geq 1+a' \left(t_{ij} - \frac{\sum_{k,l}t_{kl}}{\sum_k n_k} \right)& \\
		\ \ \ \ \ \  \forall i=1, \dots n, \ \forall j=1, \dots, n_i \\
	\end{cases}
	\label{eq.LSVM2}
\end{multline}
where
\begin{equation}
	\begin{cases}
		a = a' / \| \vv \|_2 \\
		d = \left(1-\frac{\sum_{i,j}a't_{ij} } {\sum_i n_i} \right)/ \|\vv\|_2\\
		b = b'/\| \vv \|_2 \\
		\wv = \vv / \|\vv\|_2
	\end{cases}
\end{equation}

Equation (\ref{eq.LSVM2}) is known as a quadratic programming problem with objective function expressed as a quadratic function of variable $\wv$ and linear inequality constraints. It can be solved by standard optimization packages. However, we are more interested in the dual form for the longitudinal data set as it has higher efficiency in coping with large data sets by taking the advantage of the powerful kernel trick.

The idea is to consider the Lagrangian form of eq. (\ref{eq.LSVM2}) by introducing the Lagrangian multipliers (also known as dual variables).
Thus, we take into account $\alpha_{ij} \geq 0, \forall i=1, \dots n, \ \forall j=1, \dots, n_i $ and the dual form of the eq. (\ref{eq.LSVM2}) is:
\begin{equation}
L =\frac{1}{2}\|\vv\|_2^2 - \sum_{i,j}^{n_i}\alpha_{ij} \left[y_i
\left(\vv^TX_i(t_{ij})+b' \right)-(a't_{ij}+d')\right].
\label{dual_LSVM}
\end{equation}
From $\partial L/\partial \vv = 0$, we get
\begin{equation}
\vv  = \sum_{ij}\alpha_{ij}y_iX_i(t_{ij}).
\label{dual_w}
\end{equation}
From $\partial L/\partial a' = 0$, we get
\begin{equation}
 \sum_{ij}\alpha_{ij}\left(\frac{\sum_{kl}t_{kl}}{\sum_k n_k}-t_{ij}\right) = 0.
 \label{dual_constraint}
\end{equation}
From $\partial L/\partial b'= 0$, we get
\begin{equation}
 \sum_{ij}\alpha_{ij}y_i = 0.
 \label{dual_b}
\end{equation}
Inserting eq. (\ref{dual_w}), (\ref{dual_constraint}) and (\ref{dual_b})  to eq. (\ref{dual_LSVM}),
the dual objective function w. r. t. variables $\alpha_{ij}$ is obtained. This is to be maximized along with the constraints,
\begin{eqnarray}
 &\arg\max_{\alpha_{ij}} &\left(-\frac{1}{2}\sum_{ij,kl}\alpha_{ij}y_iX_{i}(t_{ij})X_{k}(t_{kl})y_k\alpha_{kl}+\sum_{ij}\alpha_{ij}\right) \nonumber
	\\
	&\mbox{ \ s.t. \ }&
	\begin{cases}
 	\alpha_{ij} \geq 0, \ \ \forall~ i,j & \\		
	\sum_{ij}\alpha_{ij}\left(\frac{\sum_{kl} t_{kl}}{\sum_k n_k}-t_{ij}\right) = 0, &\\
	\sum_{ij}\alpha_{ij}y_i = 0. &
	\end{cases}
	\label{eq.dual_LSVM2}
\end{eqnarray}

With the optimized $\alpha_{ij}$, we can recover $\vv$ according to eq. (\ref{dual_w}).
The parameters $a'$, $b'$ and $d'$ can be found by solving the equation set: $$y_i(\vv^TX_i(t_{ij}) + b')  = 1+a' \left(t_{ij} - \frac{\sum_{k,l}t_{kl}}{\sum_k n_k} \right)$$ for which $\alpha_{ij}>0$. The original primal parameters $\wv$, $a$, $b$ and $d$ are obtained by normalizing $\vv$, $a'$, $b'$ and $d'$ w.r.t the norm of $\vv$.

It is straightforward to have the dual form of eq. (\ref{eq.mm2}) for the noise case.
The final formula are similar to eq. (\ref{eq.dual_LSVM2}) while the tuning parameter $C$ appearing as an additional constraint on the Lagrangian multipliers $\alpha_{ij}$ as $$0\leq \alpha_{ij}\leq C,~~ \forall~i,j.$$


\bibliography{References}
\bibliographystyle{IEEEtran}

\end{document}